\documentclass{article}
\usepackage{spconf,amsmath,graphicx}

\title{Discrete Independent Component Analysis  (DICA)  \\ with Belief Propagation}
\name{Francesco  A. N. Palmieri and  Amedeo Buonanno \thanks{Work partially supported by PON03PE-00185-1 - {\it C3ISR}, with Ministero dell'Istruzione dell'Universit\`a e della Ricerca;  and  PON03PE-00185-2 - {\it MAR.TE.}, with Consorzio Nazionale Interuniversitario per le Telecomunicazioni (CNIT).}}

\address{ Dipartimento di Ingegneria Industriale e della Informazione, \\ Seconda Universit\'a degli Studi di Napoli (SUN), Aversa, Italy \\ \{francesco.palmieri; amedeo.buonanno\}@unina2.it}

\begin{document}
%
\maketitle
\begin{abstract}
We apply belief propagation to a Bayesian bipartite graph composed  of discrete independent hidden variables  and  discrete visible variables. The network is the Discrete counterpart of Independent Component Analysis (DICA) and it is manipulated in a  factor graph form for inference and learning.     
A full set of simulations is reported for character images from the MNIST dataset. The results  show that the  factorial code implemented by the sources contributes to build a  good generative model for the data that can be used in various inference modes.
\end{abstract}
\section{Introduction}
\label{sec:intro}
Bi-directional information flow in belief propagation networks is becoming a very popular framework in many signal processing applications \cite{SPMag2010}\cite{Barber2012} because inference and learning can be easily manipulated with a small set of rules.  Generally  Bayesian models aim at capturing the hidden structure that may underly observed data through the assumption of a network of random variables that are only partially, or occasionally, visible \cite{Pearl1988}. 
  
Independent Component Analysis (ICA) is a popular signal processing framework in which observed data are mapped to, or generated from, independent hidden sources variables \cite{Hyvarinen2001}. The variables are typically continuous and the transformation between sources and visible variables is linear.   
ICA has been used in many applications for signal separation and for analyzing signals and images \cite{Hyvarinen2001}. ICA filters, trained on real images, seem to converge to patterns that resemble the receptive fields found in the neural  visual cortex \cite{Hyvarinen2009}. 

In this paper we explore the possibility of using the generative model of the ICA on discrete variables. The Bayesian model is constrained to a finite  number of discrete hidden sources (factorial code) that feed the visible variables, also discrete. Even if there are  computational difficulties that naturally emerge in dealing with the product space of discrete alphabets, we find that even limiting our attention to tractable small sizes, the DICA framework clearly shows some potential  in the applications, perhaps as a building block of more complex architectures.  Discrete Component Analysis (DCA) has also been discussed by Buntine et al. \cite{Buntine2006} with reference to different models.  

We reduce the DICA architecture to a Bayesian  factor graph in the so-called reduced normal form (see \cite{Palmieri2013} and reference therein) that includes only simple interconnected blocks. We experiment with belief propagation on this architecture using images extracted from the  MNIST dataset \cite{LeCun1998}. We show that the DICA network nicely converges after learning to a generative model that reproduces accurately the image set.   

In Section 2 the Bayesian model is presented and in Section 3 its  discrete version is transformed into a factor graph for belief propagation. The various modes of inference are discussed in Section 5 and learning in Section 6. The simulations for unsupervised mapping of the MNIST images are reported in Section 6 with the addition of the label variable in Section 7. The conclusions are in Sections 8.

\section{The Bayesian Model}

\begin{figure}[ht]
\centerline{\includegraphics[width=5.5cm]{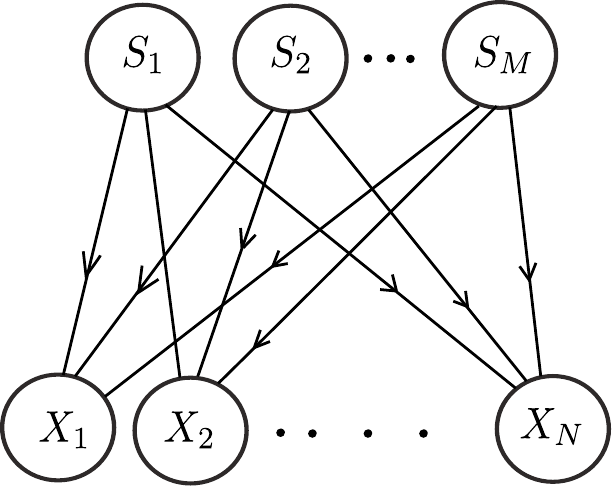}}
\caption{The Bayesian Graph for $M$ independent sources}
\label{fig:dica_B}
\end{figure}

\begin{figure}[ht]
\centerline{\includegraphics[width=5.5cm]{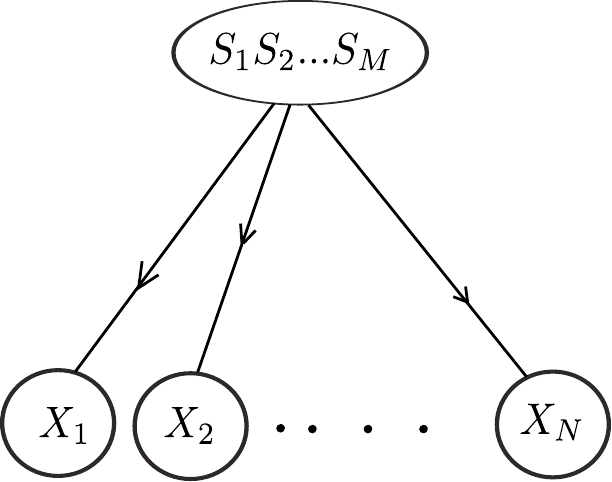}}
\caption{The Bayesian Graph for $M$ independent sources after the sources have been grouped (married).}
\label{fig:dica_BJ}
\end{figure}

In this paper we focus on the  generative model depicted as the bi-partite graph of Figure \ref{fig:dica_B} with  $M$ independent   source 
variables  $S_1,S_2,$  $....,S_M$ (hidden). The  main variables $X_1,X_2,....,X_N$ (visible),  are connected to the source variables via the factorization
\begin{equation}
\begin{array}{l}
p(X_1X_2...X_NS_1S_2...S_M)= \\
~~~~~~p(X_1|S_1S_2...S_M)p(X_2|S_1S_2...S_M)  \\ 
~~~~~~\cdot \cdot \cdot p(X_N|S_1S_2...S_M)p(S_1)p(S_2) \cdot \cdot \cdot p(S_M) \\

\end{array}
\end{equation}
Note that  $X_1,X_2,....,X_N$ to be conditionally independent, must be conditioned on the  whole set of  sources, even if their marginal distribution  factorizes: $p(S_1S_2...S_M)=p(S_1)p(S_2) \cdot \cdot \cdot p(S_M)$. This appears to be the most general model for independent hidden sources that underly a set of dependent variables  $X_1,X_2,....,X_N$. 
When $M=1$, the system degenerates into a single-variable latent model \cite{Barber2012}.  

One way of solving for the probability functions involved in the  Bayesian model is to group (marry) the  source variables (parents) \cite{Lauritzen1996} as in Figure \ref{fig:dica_BJ}. Note that the Bayesian graph does not show that the source variables are marginally independent. This is made more explicit  in the factor graph representation that will follow.

\subsection{Generative model for classical ICA} 

Independent Component Analysis is obtained when all the variables  $x_1,x_2,...,x_N,s_1,s_2,...,s_M$ $\in$ ${\cal R}$  and the conditional probability density functions $p(x_i|s_1s_2...s_M)$ are constrained to depend on linear combinations of $s_1,s_2,...,s_M$. More specifically, the  typical assumption is that the linear combinations contribute to the means of $X_1,...,X_N$ and  the dispersion around  the mean is spherical and follows a Gaussian distribution
\begin{equation}
p(x_i|s_1s_2...s_M)={\cal N}(x_i; {\bf a}_i^T {\bf s}, \sigma^2),~~~~i=1,...,N,
\end{equation} 
where the vector ${\bf s}$ contains all the source values ${\bf s}^T=[s_1 s_2 . . . s_M]$ and ${\bf a}_i$ is the $i$th column of the $N \times M$ coefficient matrix $A=[{\bf a}_1{\bf a}_2...{\bf a}_M]$  \cite{Hyvarinen2009}. More compactly $p({\bf x}|{\bf s})={\cal N}({\bf x}; A^T {\bf s}, \sigma^2 I_N)$, where ${\bf x}^T=[x_1 x_2 . . . x_N]$. The sources' pdfs $p(s_1),p(s_2),...,p(s_M)$ can follow various distributions that go from uniform to laplacian
\cite{Hyvarinen2009}.   Typically for the model to be identifiable, the sources cannot be Gaussian (except perhaps for one out of $M$). 

Unfortunately when ICA is used as a generative model it is hard to produce realistic images even when experimental densities are used as density sources \cite{Hyvarinen2009}. Structured patches are easy to obtain, but they do not resemble the complex structures found in natural images. The reason is that  independent continuous sources do not carry the  necessary structure to assemble the ICA into the complex structures found in natural images. We report a simulation in the following that seems to confirm these results.  Attempts have been made to use the ICA  in two-layer architectures \cite{Hyvarinen2009}. However, it is not clear how to properly include non linearities (without non linearities the whole system would still be linear) and investigations in this direction are still in progress.   

\begin{figure}[ht]
\centerline{\includegraphics[width=8.5cm]{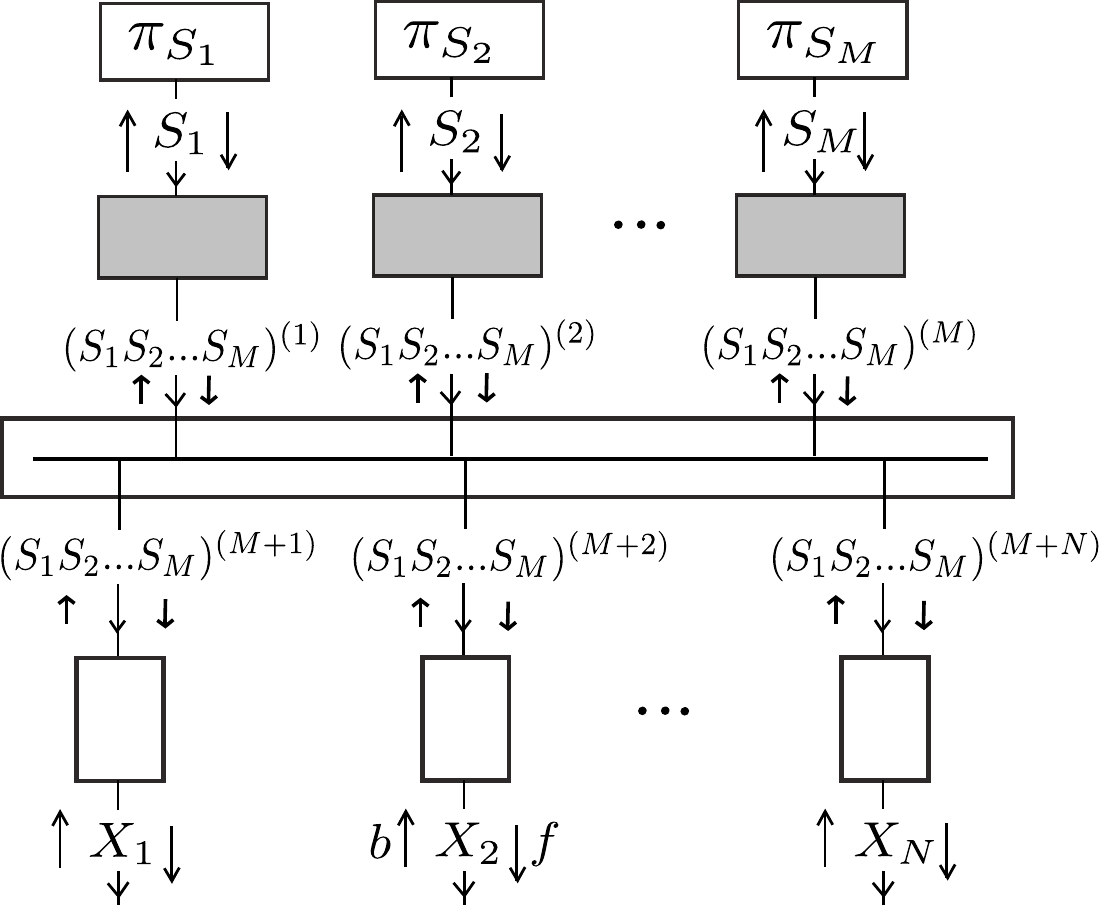}}
\caption{The DICA model as a factor graph in reduced normal form. The shaded boxes represent the fixed matrices $P(S_1S_2...S_M|S_i)$, $i=1,...,M$. The unshaded boxes represent the conditional probability matrices $P(X_j|S_1S_1...S_M)$, $j=1,...,N$.}
\label{fig:dica_FG}
\end{figure}

\subsection{Discrete  ICA}

In this work we experiment on the unconstrained ICA model with discrete variables. More specifically we assume that  both sources and visible variables take values in the finite  discrete alphabets 
${\cal S}_1,{\cal S}_2,.... $ $,{\cal S}_M$, ${\cal X}_1,{\cal X}_2,....,{\cal X}_N$, with sizes  $|{\cal S}_1|, |{\cal S}_2|,....,|{\cal S}_M|$ and $|{\cal X}_1|, |{\cal X}_2|,$ $....,|{\cal X}_N|$. 

The  difficulties in dealing with such a model are clearly related to the computational complexity in the manipulation of the product space  ${\cal S}={\cal S}_1 \times {\cal S}_2 \times ... \times {\cal S}_M$, that has size $|{\cal S}|=|{\cal S}_1||{\cal S}_2| \cdot \cdot \cdot |{\cal S}_M|$ (Figure \ref{fig:dica_BJ}). However, we find that even limiting our attention to  small dimensionalites, i.e. to few source variables and to small alphabets, the framework applied to natural images reveals quite interesting results. Furthermore, the basic architecture can be used  as a building block for more complicated multi-layer Bayesian architectures (not discussed in this paper).

\section{DICA in Reduced Normal Form}

Probability propagation and learning for the graph of Figure \ref{fig:dica_B} can be handled in a
very flexible way if we transform the model into a factor graph as in  Figure \ref{fig:dica_FG}. The graph is in the so-called {\em reduced normal} form (see \cite{Palmieri2013} and references therein), that is composed only of one-to-one blocks, source blocks and diverters (these are equal constraint blocks that act like buses for belief propagation). One-to-one  blocks are characterized by a conditional probability matrix and sources by a probability vector. We have often advocated the use of such a  representation because it can be handled as a block diagram and it is amenable to distributed  implementations. We have also designed a Simulink library for rapid prototyping \cite{Buonanno2014}.

More specifically  for the DICA model, the source variables, that have prior distributions $\Pi_{S_1}$, ... $\Pi_{S_M}$,  are mapped  to the product space via the fixed row-stochastic matrices (shaded blocks)
\begin{equation}
\begin{array}{l}
P((S_1S_2...S_M)^{(1)}|S_1) \\
~~~~~~~~~~~~~~~~~={|{\cal S}_1| \over \prod_{i=1}^M |{\cal S}_i| }I_{|{\cal S}_1|} \otimes  1_{|{\cal S}_2|}^T \otimes 1_{|{\cal S}_3|}^T \otimes ... \otimes  1_{|{\cal S}_M|}^T, \\
P((S_1S_2...S_M)^{(2)}|S_2)\\
~~~~~~~~~~~~~~~~~={|{\cal S}_2| \over \prod_{i=1}^M |{\cal S}_i| }1_{|{\cal S}_1|}^T \otimes  I_{|{\cal S}_2|} \otimes 1_{|{\cal S}_3|}^T \otimes ... \otimes  1_{|{\cal S}_M|}^T, \\ 
~~~~~~~~~~~~~~~~~..... \\
P((S_1S_2...S_M)^{(M)}|S_M)\\
~~~~~~~~~~~~~~~~~={|{\cal S}_M| \over \prod_{i=1}^M |{\cal S}_i| }1_{|{\cal S}_1|}^T \otimes  1_{|{\cal S}_2|}^T \otimes 1_{|{\cal S}_3|}^T \otimes ... \otimes  I_{|{\cal S}_M|} ,
\end{array} 
\end{equation}
where $\otimes$ denotes the Kronecker product, $1_K$ is a $K$-dimensional column vector with all ones, and $I_K$ is the $K \times K$ identity matrix.  The conditional probability matrix is such that each variable contributes to the product space with its value and it is uniform on the components that compete to the other source variables. 
The blocks at the bottom of Figure \ref{fig:dica_FG} represent the $|S| \times |{\cal X}_j|$ conditional probability matrices $P(X_j|S_1S_2...S_M)$, $j=1,...,N$, that with the source prior distributions are typically learned from data. 
Information flows in the network  bi-directionally: for each branch variable there is a forward ($f$) and a backward ($b$) message, which are (or proportional to) discrete probability vectors. Messages are usually kept normalized  for numerical stability. The variables connected to the diverter represent a replicated version of the same variable, but  they all  carry different forward and backward messages that are combined with the product rule \cite{Loeliger2004}.  Propagation through each one-to-one block follows the sum rule which in the variable direction is the matrix multiplication $f_{out}=P(out|in)^T f_{in}$ (already normalized) and in the opposite direction   $b_{in}'=P(out|in) b_{out}$ and  $b_{in}={b_{in}' \over \sum b_{in}'}$ (normalization). After propagation for a number of steps equal to the graph diameter (if there are no loops), posterior probability $p$ for a variable branch can be computed with the normalized product   
$p={f \odot b \over \sum (f \odot b)}$ ($\odot$ denotes the element-by-element product of two vectors). For the reader not familiar with this framework, it should be emphasized that these simple rules are rigorous translation of marginalization and Bayes' theorem \cite{Loeliger2004}.  

\section{Inference in the DICA graph}

The flexibility of this framework  allows the use of the factor graph of Figure \ref{fig:dica_FG} in various  inference modes. Information flow is bi-directional and assuming that all the parameters have been learned and that the unspecified messages are initialized to uniform distributions, we can use the DICA graph in:

\noindent
{\em (1) Generation:} Source values are picked and are injected as forward  delta distributions at $S_1,S_2,...,S_M$. After three steps of message propagation, the forward distributions are collected at the terminal variables
$X_1,X_2,...,X_N$. They are the (soft) decoded version of the source values.  Note that these are distributions that are typically displayed as their means or their argmaxes (see simulation results in the following).     

\noindent
{\em (2) Encoding:} Observed values for $X_1,X_2,...,X_N$ are injected as delta backward distributions at the bottom. After three steps of message propagation, the backward distributions  are multiplied with the forward at  $S_1,S_2,...,S_M$. The normalized result is a (soft) {\em factorial code of the input}. The set of argmaxes of these distribution is the MAP decoding of the input.    

\noindent
{\em (3) Pattern completion:} Only a subset of values for $X_1,X_2,...$ $,X_N$ is available (there are erasures). 
The available values are injected at the bottom as delta backward distributions. For the missing values uniform densities are usually injected. After three  steps of message propagation,  forward distributions are collected at the bottom variables. For the observed variables the forward-backward products return just the deltas on the observations and provides no new information. At the unknown variables, the forward distribution is our best (soft) knowledge of that variable. Here too the means or the argmaxes can be used as a final result. The inference on the erasures is the synthesis of the information coming from the  observations and the priors. 

\noindent
{\em (4) Error correction:} Available values for $X_1,X_2,...,X_N$ may contain errors. They are presented as backward delta distributions at the bottom variables. After three  steps of message propagation,  forward distributions (or their means or argmaxes) are collected and used as  corrections. No product with the backward is applied here because we do not know which component is reliable. In a similar scheme the values for $X_1,X_2,...,X_N$ may be known softly via distributions that are injected at the bottom as backward messages.     

Note that in both {\em (3)} and {\em (4)} also coded versions of the observations are available at the source branches. 

\section{Learning in the DICA graph}

 To train the DICA system, we assume that a set of T examples is available for the visible variables $(x_1[n]x_2[n]...x_N[n]), n=1,...,T$ (training set). Learning the system matrices for the bottom blocks and the vectors for the sources, is performed using an EM search. Various algorithms can be used, all inspired by a localized  maximum likelihood cost function. The iterations are  confined to each block and use only locally available forward and backward messages. Details on the learning algorithms for the factor graph in reduced normal form have been reported elsewhere and are omitted here for space reasons (see \cite{PalmieriTSP2013} \cite{Palmieri2013} and references therein).

\begin{figure}[ht]
\centerline{\includegraphics[width=1\linewidth]{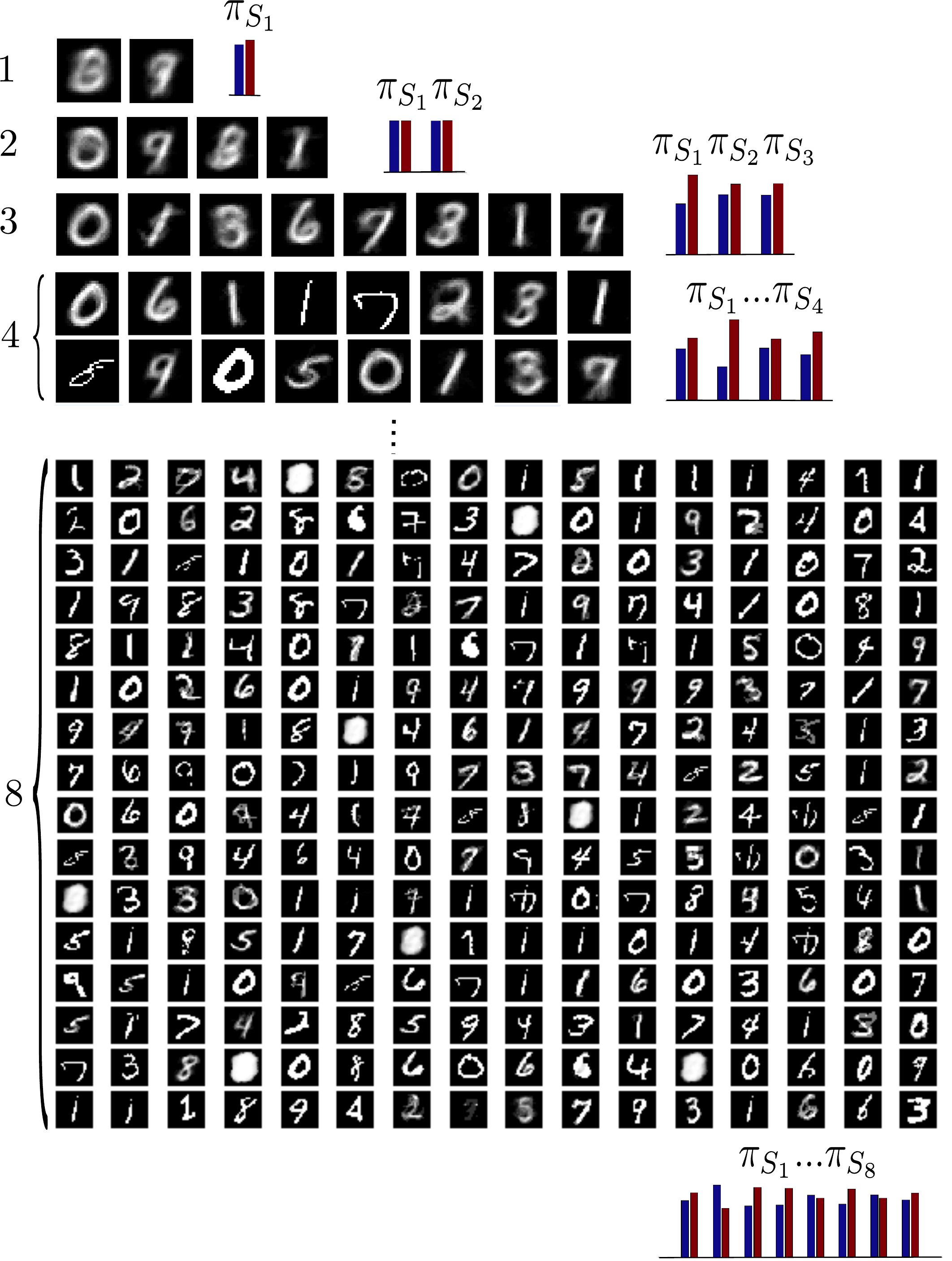}}
\caption{Distribution means generated by the factorial code for increasing number of sources ($M=1,2,3,4,8$). The bars show the learned source priors.} 
\label{fig:IncreasingSources}
\end{figure}

\section{DICA Simulations}

We report here a full set of simulations on the MNIST data set \cite{LeCun1998}. We have reduced the images to $ 28 \times 28 $ binary pixels and extracted  500 images as our training set.  
In a first set of experiments we train the architecture of Figure \ref{fig:dica_FG} with all binary variables: ${\cal X}_j=\{x^0,x^1\}$, $j=1,...,N$ ($N=784$);
 ${\cal S}_i=\{s^0,s^1\}$, $i=1,...,M$, for various number of sources $M=1,2,3,4,8$. 
During learning the 500 images of the training set are presented as backward delta distributions on $X_1,...,X_N$,  one time, with 5 cycles inside each block (the maximum likelihood algorithm inside each block is iterative \cite{Palmieri2013}).  Therefore for each order $M$ we obtain the conditional probability matrices $P(X_j|S_1...S_M)$, $j=1,...,N$, and the prior distributions $\pi_{S_1},...,\pi_{S_M}$. 

\noindent
{\em Generation:} Figure \ref{fig:IncreasingSources} shows, for increasing $M$, the means of $f_{X_1},...,f_{X_N}$ when at the sources we inject the $2^M$ binary configurations in the forward messages $f_{S_1},...,f_{S_M}$. Reported in the picture are also the learned priors. We note that, for larger number of sources, the product space (sizes 2,4,8,16,256), corresponds to increasingly accurate pattern memorization. For some characters, that are different in shape, the system builds separate representations. The source variables, independent by definition (factorial code), learn marginal distributions progressively less uniform as the number of sources increases (recall that the vector that represents $p(S_1,....,S_M)$ is the Kronecker product of the individual binary distributions and that even small non uniformities in the priors cause $p(S_1,....,S_M)$ to be highly non uniform).  

\begin{figure}[ht]
\centerline{\includegraphics[width=0.6\linewidth]{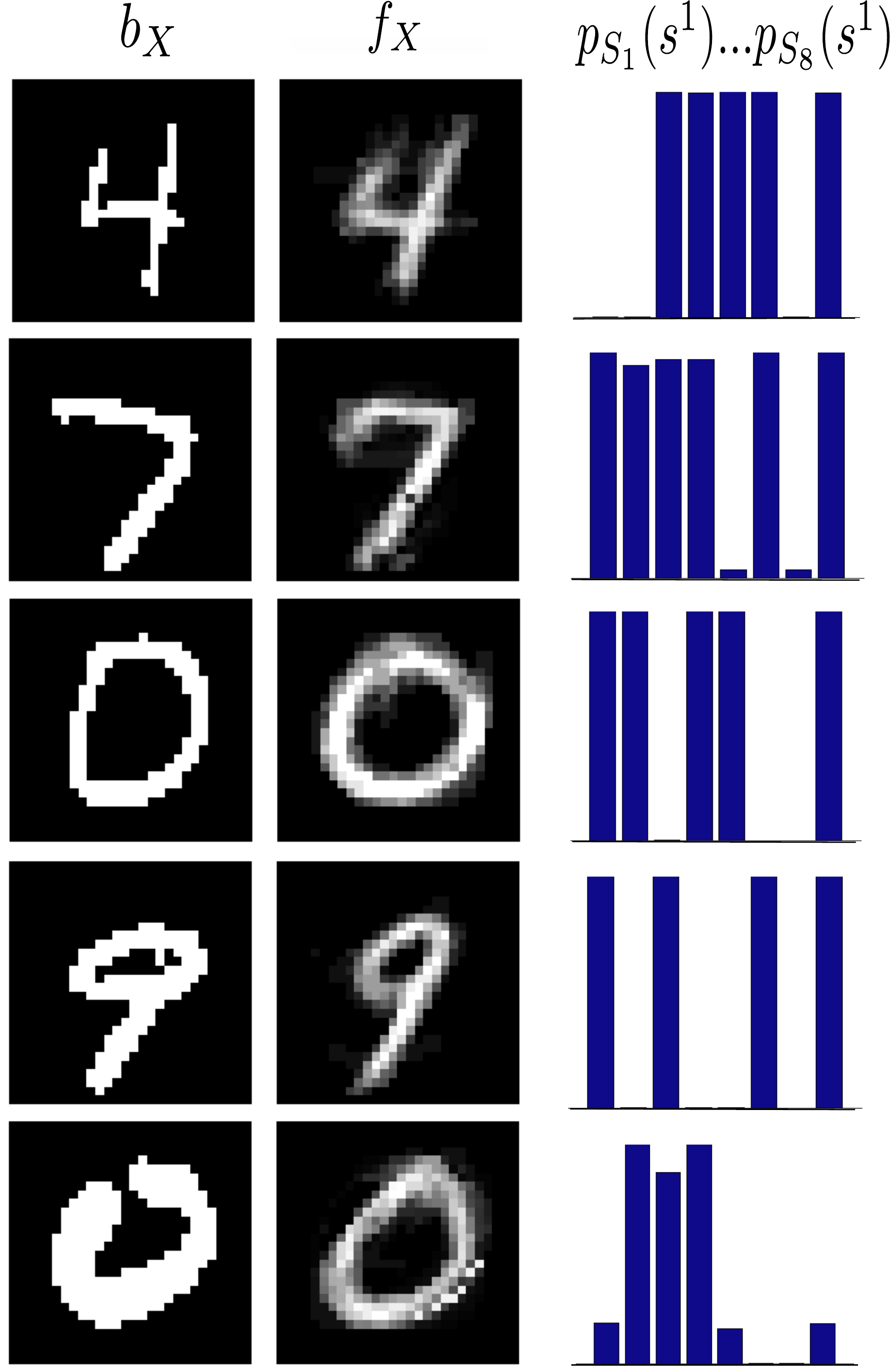}}
\caption{Encoding of some images from the test set. Col. 1: images presented as delta backward distributions. Col. 2: means of the forward distributions. Col. 3:  posterior probabilities at the sources (the bars represent $[p_{S_1}(s^1)...p_{S_8}(s^1)]$).}
\label{fig:Encoding}
\end{figure}

\noindent
{\em Encoding}: Figure \ref{fig:Encoding} shows the typical results of presenting to the DICA graph  of Figure \ref{fig:dica_FG}, with $M=8$, images from the test set (i.e. not included in the 500 images used for training) as backward delta distributions at $X_1,...,X_N$. In the third column the posterior distributions at the sources are shown (only the  probability on the symbol $s^1$ is depicted). Here the DICA graph acts as an Encoder: the (soft) binary configurations are the factorial code of the presented images. Note that not all the codes are sharp.  In the second column the mean of the forward distributions at $X_1,...,X_N$ is also shown.  

\begin{figure}[ht]
\centerline{\includegraphics[width=0.8\linewidth]{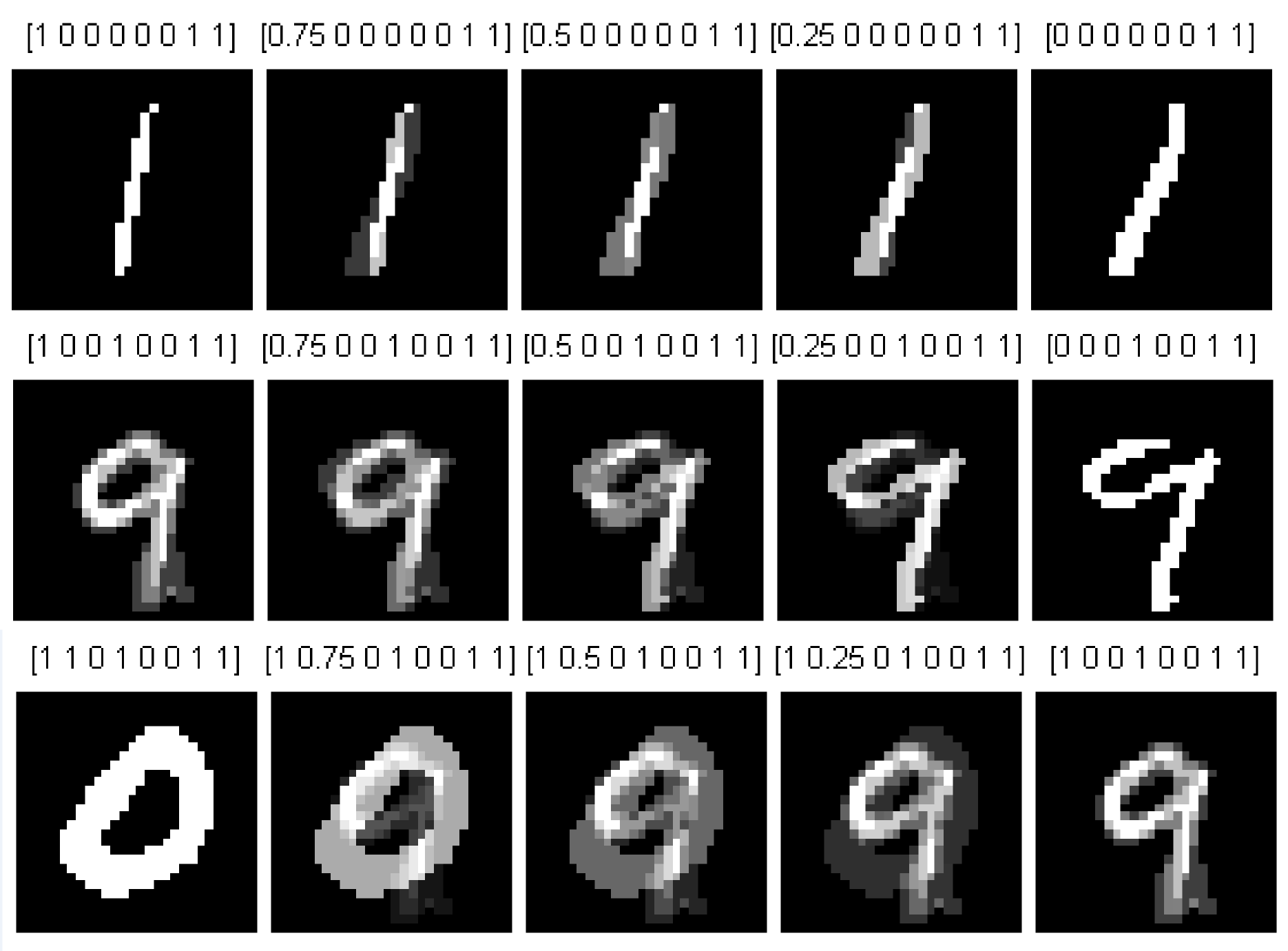}}
\caption{Decoding for smooth forward distributions at the sources (in the brackets the probabilities $[f_{S_1}(s^1)...f_{S_8}(s^1)]$)}
\label{fig:Decoding}
\end{figure}

\noindent
{\em Decoding}: In Figure \ref{fig:Decoding} the same DICA graph is used as a soft decoder when smooth and sharp distributions are injected at the sources. 

\begin{figure}[ht]
\centerline{\includegraphics[width=0.6\linewidth]{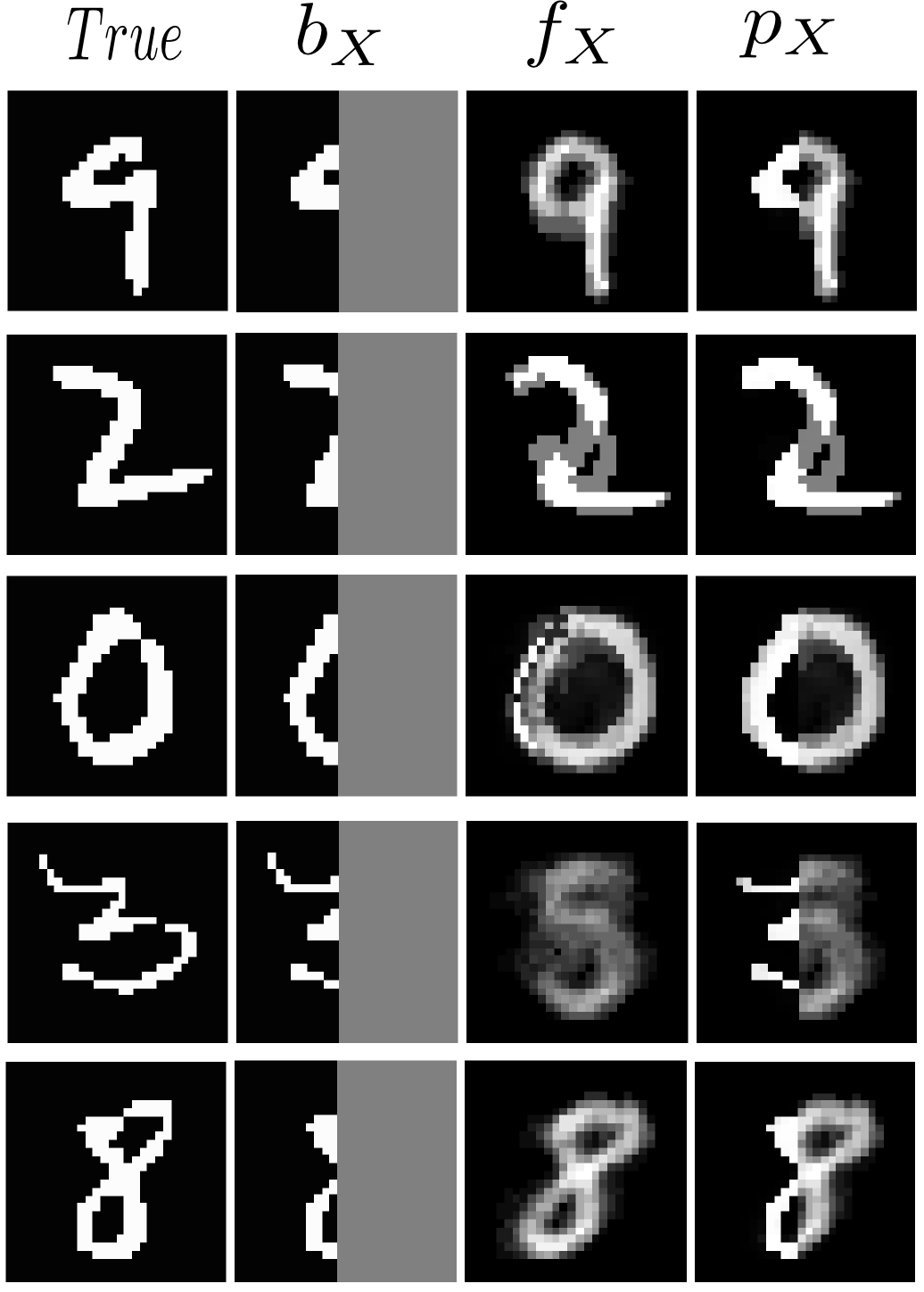}}
\caption{Pattern completion of images from the test set after 50\% removal.}
\label{fig:Recall}
\end{figure}

\noindent
{\em Pattern completion}: Figure \ref{fig:Recall} shows the results of the same network when as backward at $X_1,...,X_N$ we present images (from the test set) with 50 \% of the pixels removed.
For the erased pixels a backward uniform distribution is presented. The third and the fourth columns report the mean for the forward and the posterior distributions respectively. The network  fills-in rather well the missing parts. 

\begin{figure}[h]
\centerline{\includegraphics[width=1\linewidth]{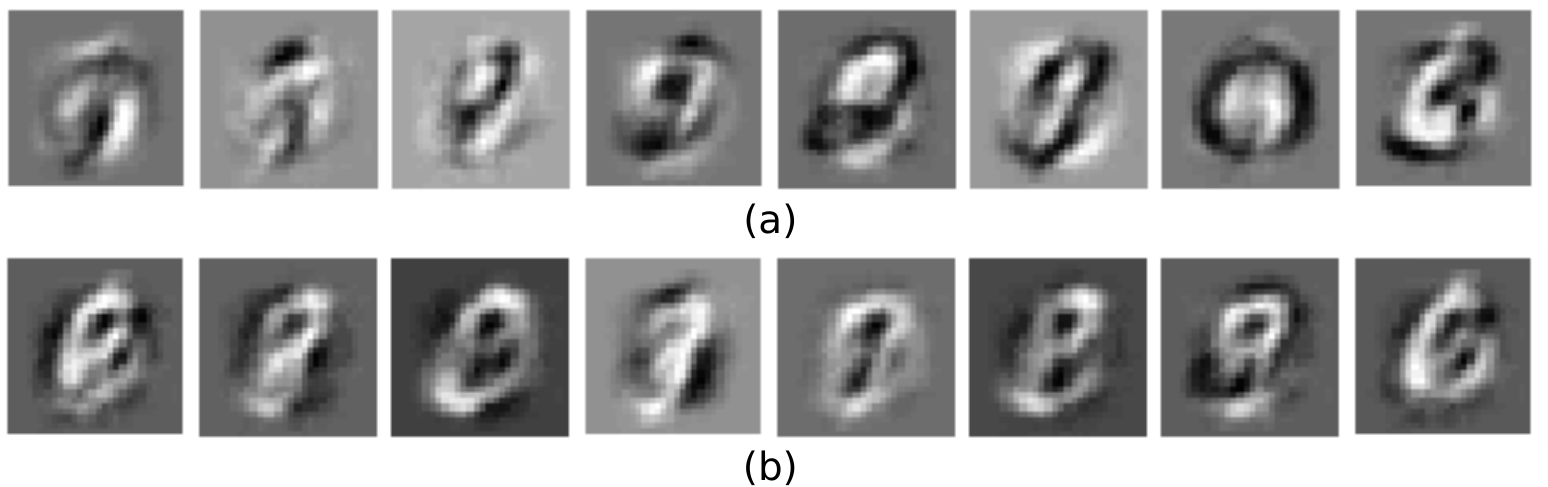}}
\caption{Continuous ICA comparison: (a) 8 ICA masks for the Training Set (b) 8 generated images using at the sources random values drawn from estimated output histograms. }
\label{fig:ICA}
\end{figure}

\subsection{Continuous ICA on the same dataset}

The natural question at this point is whether with continuous ICAs it would be possible to obtain similar results. The model is clearly very different, but on the same data set we have attempted a comparison. On the 500 MNIST images of the training set we have computed ICAs using the  Fast ICA algorithm available for Matlab \cite{FastICAMatlab}. We have retained only the first 8 components  (largest variance) and estimated the output densities using average histograms. Random samples from these densities are used to generate the images though the inverse ICA \cite{Hyvarinen2010}. Figure \ref{fig:ICA} shows the 8 masks and some generated images. The results confirm that, even if the ICA nicely represent bases for the data, with unconstrained independent samples at the sources, only average structures are generated. We have also tried with larger number of components and the obtained images look very similar. These results seem to be  consistent  with other experiments  presented in the literature \cite{Hyvarinen2010} for patches of natural images where only average textures are obtained. The linear ICA with independent unconstrained sources do not seem to be a generative model that preserves  the structured composition of  the training set.

\section{DICA for classification}

\begin{figure}[ht]
\centerline{\includegraphics[width=6.5cm]{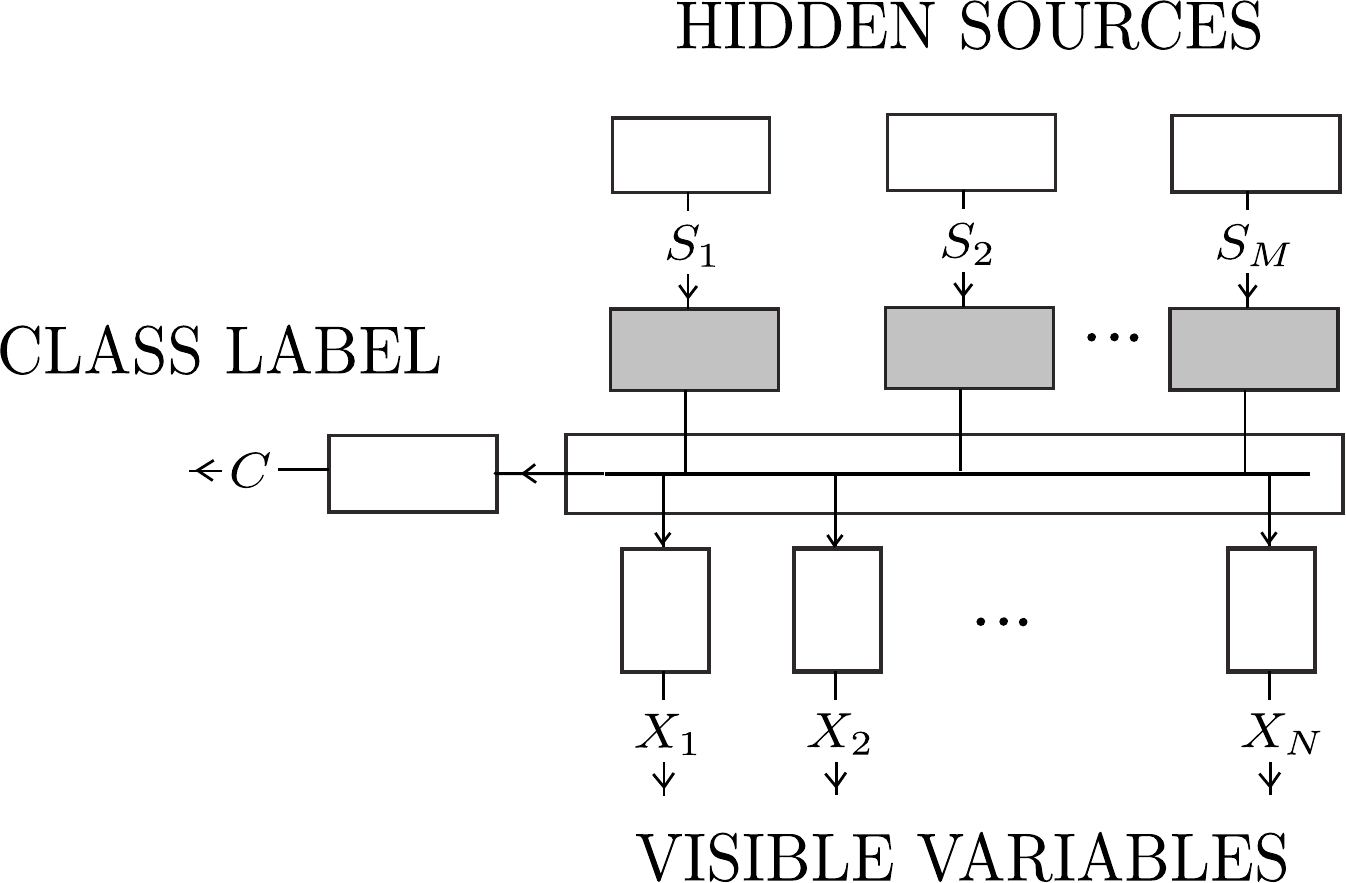}}
\caption{The DICA model for classification}
\label{fig:dica_FGL}
\end{figure}

The great flexibility of the factor graph framework allows to extend easily the architecture of the DICA graph to the one shown in Figure \ref{fig:dica_FGL} where also a label variable $C$ is included. 
The variable $C$ belong to the finite alphabet ${\cal C}={c^0,c^1,...,c^9}$ and it is attached directly, through a conditional probability matrix $P(C|S_1,...,S_M)$, to the product space diverter. Diverters in the reduced normal form act like probability pipelines \cite{Palmieri2013}. 

\begin{figure}[ht]
\centerline{\includegraphics[width=1\linewidth]{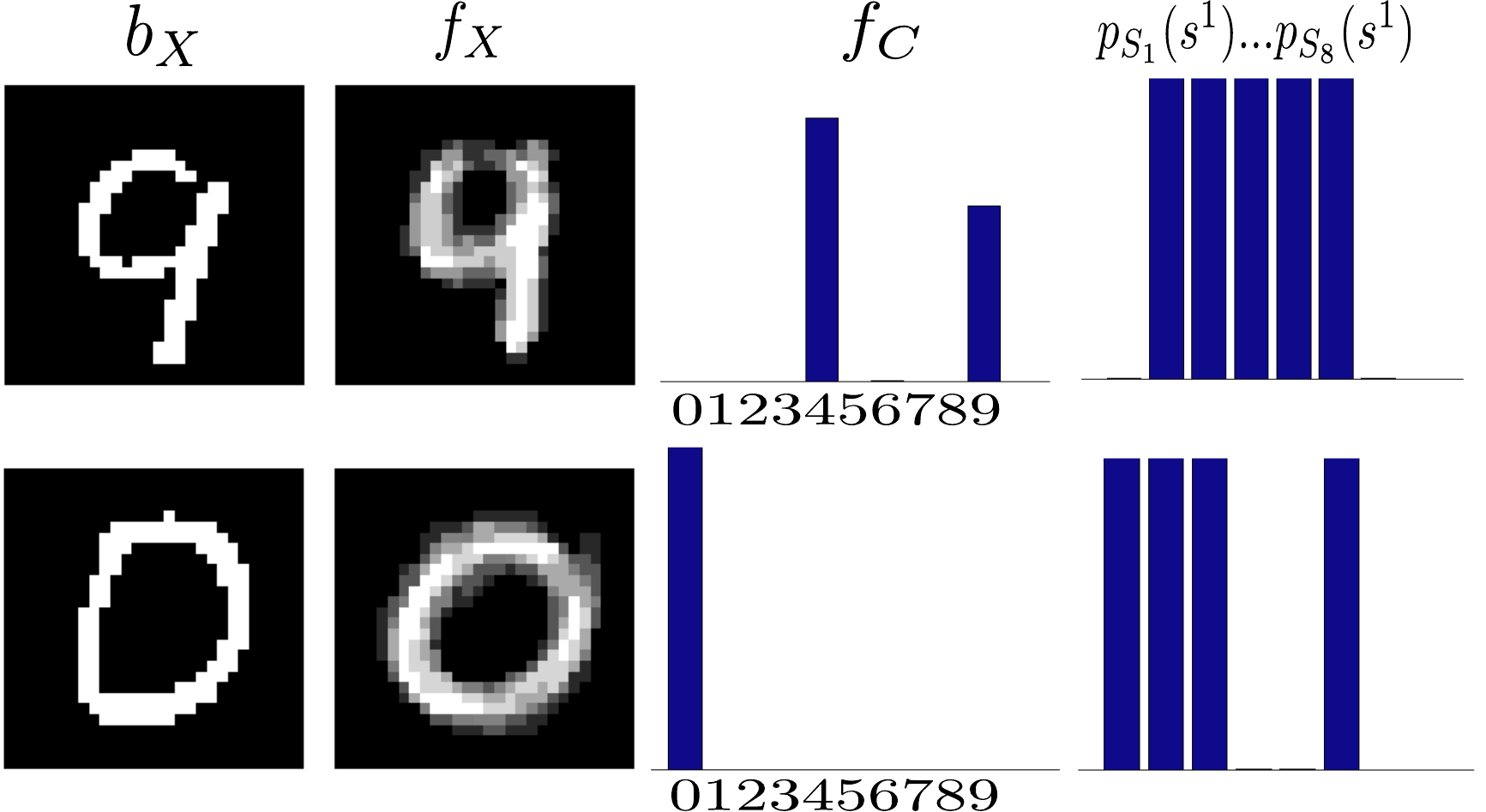}}
\caption{Recognition task on two images from the test set}
\label{fig:Recognition}
\end{figure}

Simulations have been performed on the same MNIST training set of 500 binarized images in the same mode as in the unsupervised experiments with the addition, during training, of the label information as a backward delta distribution. All the blocks, including now the  probability matrix $P(C|S_1,...,S_M)$, are trained for $M=8$. On the learned  network, a typical recognition task on two images from the test set is shown in Figure \ref{fig:Recognition}. The bar graph represents simultaneously classification and encoding. Note how in the first row the network is naturally confused between $c^4$ and $c^9$. 

\begin{figure}[ht]
\centerline{\includegraphics[width=1\linewidth]{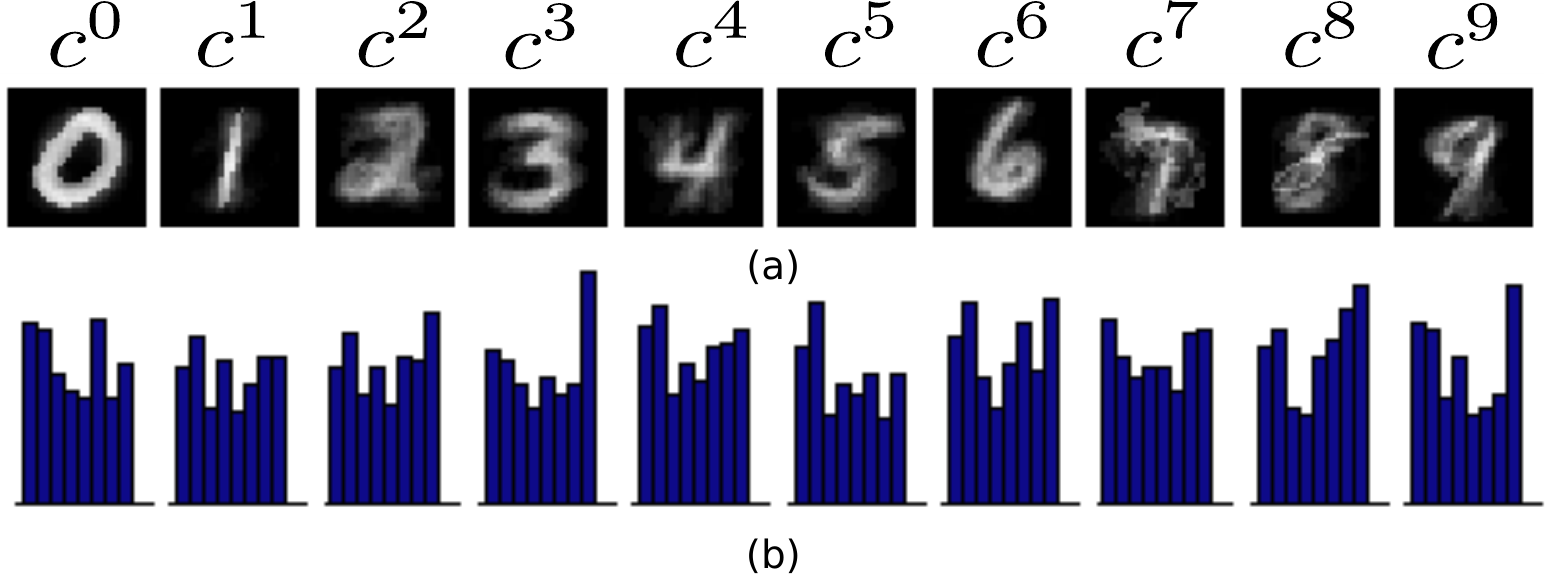}}
\caption{Results of injecting  backward deltas in $C$, $b_{C}(c)=\delta(c,c^i)$, $i=0,...,9$. (a) Means of the forward distributions (prototypes); (b) Posterior probabilities at the sources $[p_{S_1}(s^1)...p_{S_8}(s^1)]$ (encoding).}
\label{fig:Prototypes}
\end{figure}

A generative experiment is also performed on this architecture with backward delta distributions injected at $C$. The results are shown in Figure \ref{fig:Prototypes}. The images are the mean forward distributions at $X_1,...,X_N$ and could be considered as the {\it prototypes} for the ten labels. The bar graphs are the corresponding simultaneous encoding at the sources.  
 
\section{Conclusions}
\label{sec:conc}
 
The simulations on the MNIST dataset with binary sources show that belief propagation in the DICA architecture, also with the addition of the label variable, provides a unified framework in which image data can be coded, generated and corrected in a very flexible way. We have also experimented on natural images on quantized patches obtaining very similar results, also when the sources have alphabet sizes greater than two. These results will be reported elsewhere. We are currently pursuing the use of this framework for building multi-layer architectures.

\small

\bibliographystyle{IEEEbib}
\bibliography{ProbPropBib}

\end{document}